# A Constraint Satisfaction Approach to the Robust Spanning Tree Problem with Interval Data

**Ionut Aron**
Computer Science Dept.
Brown University
Box 1910, Providence, RI 02912

**Pascal Van Hentenryck**
Computer Science Dept.
Brown University
Box 1910, Providence, RI 02912

## Abstract

Robust optimization is one of the fundamental approaches to deal with uncertainty in combinatorial optimization. This paper considers the robust spanning tree problem with interval data, which arises in a variety of telecommunication applications. It proposes a constraint satisfaction approach using a combinatorial lower bound, a pruning component that removes infeasible and suboptimal edges, as well as a search strategy exploring the most uncertain edges first. The resulting algorithm is shown to produce very dramatic improvements over the mathematical programming approach of Yaman et al. and to enlarge considerably the class of problems amenable to effective solutions.

## 1 Introduction

In many combinatorial optimization problems, data are not known with certainty. How to deal with these uncertainties is often problem-dependent and various frameworks have been proposed in the past, including stochastic programming [Birge and Louveaux, 1997], stochastic constraint programming [Walsh, 2001], and robust optimization [Kouvelis and Yu, 1997] to name only a few. In robust optimization, the idea is to find a solution that hedges against the worst possible scenario. More precisely, the goal is to compute a solution that minimizes the maximum deviation from the optimal solution over all realizations of the random data. Many practical applications can be formulated as robust optimization problems.

This paper considers the robust spanning tree problem in graphs where the edge costs are given by intervals (RSTPIE). The goal in the RSTPIE is to find a spanning tree that minimizes the maximum deviation of its cost from the minimum spanning tree (MST) over all possible realizations of the edge costs (i.e., costs within the given intervals). The robust spanning tree was studied in [Bertsekas and Gallagher, 1987, Kouvelis and Yu, 1997, Kozina and Perepelista, 1994, Yaman et al., 2001] because of its importance in communication networks. [Kouvelis and Yu, 1997] discusses the design of communication networks where the routing delays on the edges are uncertain, since they depend on the network traffic. A robust spanning tree hedges against the worst possible delays and is desirable in this problem. A second application discussed by [Bertsekas and Gallagher, 1987] concerns broadcasting of messages from a given node to all other nodes in a network where the transmission times are uncertain. Once again, the goal is to hedge the broadcast against the worst communication delays.

Recently, a very elegant mixed integer programming (MIP) approach was proposed in [Yaman et al., 2001]. It combines the single commodity model of the minimum spanning tree with the dual of the multicommodity model for the same problem. In addition, they introduced the concepts of weak and strong edges, which identify edges that may and must belong to an optimal solution. They used these concepts as a part of preprocessing step and showed that, on some classes of instances, it significantly enhance the performance of their MIP implementation.

This paper applies a constraint satisfaction approach to the RSTPIE. It presents a search algorithm based on three important and orthogonal components: (1) a *combinatorial lower bound* to eliminate suboptimal trees; (2) a *pruning component* which eliminates edges that cannot be part of any feasible or optimal solution; and (3) a *branching heuristic* which explores the most uncertain edges first. The lower bound exploits the combinatorial structure of the problem and reduces to solving two MST problems. The pruning component eliminates infeasible edges (e.g., edges that would lead to cycles and non-connected components) as well as suboptimal edges (i.e., edges that cannot appear in op-timal solutions). It uses the concept of weak edges and a new $O(m \log m)$ amortized algorithm for detecting all weak edges (where $m$ is the number of edges in the graph), improving the results of [Yaman et al., 2001] by an order of magnitude.

The constraint satisfaction algorithm is shown to provide very dramatic speed-ups over the MIP approach. In particular, it runs several hundred times faster than the MIP approach on the instance data proposed in [Yaman et al., 2001]. It also exhibits even more significant speed-ups on other instances which have more structure. In addition, the constraint satisfaction approach significantly broadens the class of instances that are amenable to effective solutions. Observe also that the constraint satifaction approach should apply equally well to other robust optimization problems, such as robust matching and robust shortest paths, which also arise in telecommunications and computational finance. Combinatorial lower bounds can be obtained similarly and the challenge is mainly to find effective characterizations of suboptimal solutions. As a consequence, we believe that constraint satisfaction approaches are likely to play a fundamental role in the solving of robust optimization problems in the future. We also believe that the concept of suboptimality pruning, i.e., removing values that cannot appear in any optimal solution, is fundamental and deserves further study for optimization problems in general.

The rest of the paper is organized as follows. Sections 2 and 3 define the problem and discuss prior work. Sections 4, 5, 6, and 7 discuss the search algorithm, the lower bound, suboptimality pruning, and branching. Section 8 presents the experimental results, and Section 9 concludes the paper.

## 2  The Problem

Informally speaking, the robust spanning tree problem, given an undirected graph with interval edge costs, amounts to finding a tree whose cost is as close as possible to that of a minimum spanning tree under any possible assignment of costs. This section defines the problem formally and introduces the main concepts and notations used in the paper.

We are given an undirected graph $G = (V, E)$ with $|V| = n$ nodes and $|E| = m$ edges and an interval $[\underline{c_e}, \overline{c_e}]$ for the cost of each edge $e \in E$. A **scenario** $s$ is a particular assignment of a cost $c_e \in [\underline{c_e}, \overline{c_e}]$ to each edge $e \in E$. We use $c_e^s$ to denote the cost of edge $e$ under a given scenario $s$.

Recall that a *spanning tree* for $G = (V, E)$ is a set of edges $T \subseteq E$ such that the subgraph $G' = (V, T)$ is acyclic and connected. The cost of a spanning tree $T$ for a scenario $s$, denoted by $c_T^s$, is the sum of the costs of all edges under scenario $s$:

$$c_T^s = \sum_{e \in T} c_e^s.$$

A **minimum spanning tree** for scenario $s$, denoted by $MST^s$, is a spanning tree with minimal cost for scenario $s$ and its cost is denoted by $c_{MST^s}$. Following [Yaman et al., 2001], we now define the worst case scenario for a spanning tree $T$ and the robust deviation of $T$, two fundamental concepts for this paper.

**Definition 1 (Relative worst case scenario).** *Given a spanning tree $T$, a scenario $w(T)$ which maximizes the difference between the cost of $T$ and the cost of a minimum spanning tree under $w(T)$ is called a relative worst case scenario for $T$. More precisely, a relative worst case scenario $w(T)$ satisfies*

$$w(T) \in \text{arg-}\max_{s \in S} (c_T^s - c_{MST^s}) \quad (1)$$

*where $S$ is the set of all possible scenarios.*

Note that $\text{arg-}\max_{s \in S} f(s)$ denotes the set

$$\{e \in S \mid f(e) = \max_{s \in S} f(s)\}.$$

and arg-min is defined similarly.

**Definition 2 (Robust Deviation).** *The robust deviation of a spanning tree $T$, denoted by $\Delta_T$, is the distance between the cost of $T$ and the cost of a minimum spanning tree under the relative worst case scenario of $T$:*

$$\Delta_T = c_T^{w(T)} - c_{MST^{w(T)}}. \quad (2)$$

The goal of this paper is to compute a robust spanning tree, i.e., a spanning tree whose robust deviation is minimal.

**Definition 3 (Robust Spanning Tree).** *A (relative) robust spanning tree is a spanning tree $T^*$ whose robust deviation is minimal, i.e.,*

$$T^* \in \text{arg-}\min_{T \in \Gamma_G} \max_{s \in S} (c_T^s - c_{MST^s})$$

*where $\Gamma_G$ is the set of all spanning trees of $G$ and $S$ is the set of all possible scenarios. According to the definition of $w(T)$ (Eq. 1), this is equivalent to:*

$$T^* \in \text{arg-}\min_{T \in \Gamma_G} (c_T^{w(T)} - c_{MST^{w(T)}}).$$

Figure 1 depicts these concepts graphically. The horizontal axis depicts various scenarios. For each scenario $s$, it shows the cost $c_T^s$ of $T$ under $s$ and the cost $c_{MST^s}$ of an MST under $s$. In the figure, scenario $s_5$ is a worst-case scenario, since the distance $c_T^{s_5} - c_{MST^{s_5}}$ is maximal and this distance is then the robust deviation of $T$. Such a figure can be drawn for each spanning tree and we are interested in finding the spanning tree with the smallest robust deviation. [Kouvelis and Yu, 1997] conjectured that the robust spanning tree problem with interval edges is NP-complete.

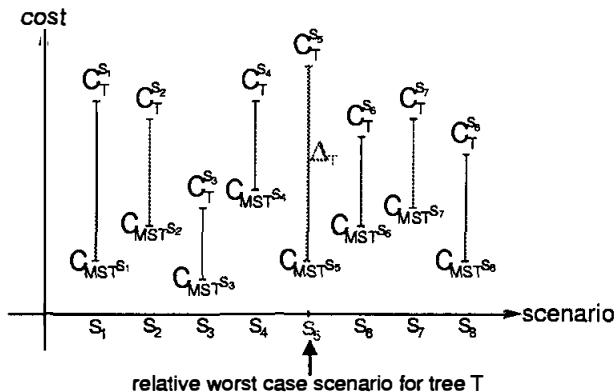

Figure 1: The relative worst case scenario determines the robust deviation $\Delta_T$ of a given spanning tree $T$. Our goal is to find a tree $T^*$ for which this deviation (distance) is minimal.

## 3  Prior Work

[Yaman et al., 2001] proposed an elegant MIP formulation for the RSTPIE problem. The formulation combines the single commodity model of the minimum spanning tree with the dual of the multicommodity model for the same problem. In addition, they introduced the concept of weak and strong edges and used them for preprocessing. They showed that preprocessing significantly enhances the performance of their MIP implementation. We now summarize their relevant results.

We first introduce the concept of weak edges, i.e., the only edges that need to be considered during the search.

**Definition 4 (Weak Tree).** *A tree $T \subseteq E$ is weak if there exists at least one scenario under which $T$ is a minimum spanning tree of $G$.*

**Definition 5 (Weak Edge).** *An edge $e \in E$ is weak if it lies on at least one weak tree.*

Strong edges are edges that are necessarily part of a robust spanning tree.

**Definition 6 (Strong Edge).** *An edge $e \in E$ is strong if it lies on a minimum spanning tree of $G$ for all possible scenarios.*

The following propositions characterize weak and strong edges in terms of the minimum spanning trees of two scenarios.

**Proposition 1.** *An edge $e \in E$ is weak if and only if there exists a minimum spanning tree using edge $e$ when its cost is at the lowest bound and the costs of the remaining edges are at their highest bounds.*

**Proposition 2.** *An edge $e \in E$ is strong if and only if there exists a minimum spanning tree using edge $e$ when its cost is at the highest bound and the costs of the remaining edges are at their lowest bounds.*

As a consequence, [Yaman et al., 2001] showed that it is possible to use a slightly modified version of Kruskal's algorithm to find all the weak and strong edges of a given graph in $O(m^2 \log m)$ time. In Section 6, we give an improved algorithm for finding weak edges, which runs in $O(m \log m)$. The following two propositions capture the intuition we gave earlier on weak and strong edges.

**Proposition 3.** *A relative robust spanning tree $T$ is a weak tree. Thus an edge can be part of a relative robust spanning tree only if it is a weak edge.*

**Proposition 4.** *There exists a relative robust tree $T$ such that every strong edge in the graph lies on $T$.*

The next result is also fundamental: it makes it possible to characterize precisely the worst case scenario of a spanning tree $T$.

**Proposition 5.** *The scenario in which the costs of all edges in a spanning tree $T$ are at upper bounds and the costs of all other edges are at lower bounds is a relative worst case scenario for $T$. In other words, $w(T)$ is specified as*

$$c_e^{w(T)} = \begin{cases} \overline{c_e}, & e \in T \\ \underline{c_e}, & e \in E \setminus T \end{cases} \quad (3)$$

In other words, once we select a tree $T$, we can easily find the worst-case scenario for $T$ by assigning to the edges in $T$ their upper bounds and to the edges not in $T$ their lower bounds. We use Proposition 5 in our new algorithm. In the rest of the paper, we also use the notation $w(S)$ to denote the scenario where $c_e = \overline{c_e}$ if $e \in S$ and $c_e = \underline{c_e}$ otherwise even when $S$ is not a tree.

## 4  The Search Algorithm

Figure 2 gives a high-level description of the search algorithm. A node in the search tree is called a configuration to avoid confusion with the nodes of the graph. A configuration is a pair $\langle S, R \rangle$, where $S$ represents the set of selected edges and $R$ represents the set of rejected edges. The algorithm receives a configuration

```
procedure Search(⟨S, R⟩)
begin
  if |S| < n - 1 then
    ⟨S, R⟩ = PruneInfeasible(⟨S, R⟩);
    ⟨S, R⟩ = PruneSuboptimal(⟨S, R⟩);
    if LB(⟨S, R⟩) ≤ f* then
      e = SelectEdge(E \ (S ∪ R));
      Search(⟨S, R ∪ {e}⟩);
      Search(⟨S ∪ {e}, R⟩);
  else
    if Δ_S < f* then
      T* = S; f* = Δ_S
end
```

Figure 2: The Search Algorithm

| Notation | Definition |
|---|---|
| $c_T^s$ | the cost of tree $T$ under scenario $s$ $c_T^s = \sum_{e \in T} c_e^s$ |
| $w(T)$ | worst case scenario for edge set $T$: $c_e^{w(T)} = \begin{cases} \overline{c_e}, & e \in T \\ \underline{c_e}, & e \in E \setminus T \end{cases}$ |
| $\Gamma_G$ | the set of all spanning trees of graph $G$ |
| $\mathcal{T}(S, R)$ | the set of spanning trees derived from $(S, R)$ $\mathcal{T}(S, R) = \{T \in \Gamma_G | S \subseteq T \subseteq E \setminus R\}$ |
| $\mathcal{M}^s(S, R)$ | the set of spanning trees from $\mathcal{T}(S, R)$ with minimal cost under scenario $s$ $\mathcal{M}^s(S, R) = \arg\text{-}\min_{T \in \mathcal{T}(S,R)} c_T^s$ |
| $MST^s(S, R)$ | an arbitrary tree from $\mathcal{M}^s(S, R)$ |
| $\mathcal{M}^s$ | $\mathcal{M}^s(\emptyset, \emptyset)$ |
| $MST^s$ | an arbitrary tree from $\mathcal{M}^s(\emptyset, \emptyset)$ |

Table 1: Summary of Notations

as input. If the configuration is not a spanning tree, the algorithm prunes infeasible and suboptimal edges. Both steps remove edges from $E \setminus (S \cup R)$ and adds them to $R$. If the lower bound of the resulting configuration is smaller than the best found solution, the algorithm selects an edge and explores two subproblems recursively. The subproblems respectively reject and select the edge. The best found solution and upper bound are updated each time a spanning tree with a smaller robust deviation is obtained.

Infeasibility pruning is relatively simple in our algorithm. It ensures that $S$ can be extended into a spanning tree and removes edges that would create cycles. We do not discuss infeasibility further in this paper and focus on the lower bound and suboptimality pruning, i.e. the key novelty.

Before doing so, we introduce some additional notations. We use $\mathcal{T}(S, R)$ to denote the set of all spanning trees that can be derived from configuration $\langle S, R \rangle$, i.e.,

$$\mathcal{T}(S, R) = \{T \in \Gamma_G \mid S \subseteq T \subseteq E \setminus R\}.$$

Given a scenario $s$, $\mathcal{M}^s(S, R)$ denotes the set of all minimum spanning trees which are derived from configuration $\langle S, R \rangle$ under scenario $s$ and $MST^s(S, R)$ is a representative of $\mathcal{M}^s(S, R)$. For simplicity, we use $\mathcal{M}^s$ and $MST^s$ when $S = \emptyset$ and $R = \emptyset$. All relevant notations are summarized in Table 1 for convenience.

## 5 The Lower Bound

We now present a lower bound to the robust deviation of any spanning tree derived from $\langle S, R \rangle$. In other words, we need to find a lower bound on the value of $\Delta_T$ (as defined by Eq. 2), for any tree $T \in \mathcal{T}(S, R)$. Recall that, for any such tree $T$, the robust deviation is given by $\Delta_T = c_T^{w(T)} - c_{MST^{w(T)}}$. We can approximate $\Delta_T$ by finding a lower bound to $c_T^{w(T)}$ and an upper bound to $c_{MST^{w(T)}}$. Since $S \subseteq T \subseteq S \cup L$ where $L = E \setminus (S \cup R)$, both bounds can be obtained by considering the scenario $w(S \cup L)$. As a consequence, we define the lower bound $LB(\langle S, R \rangle)$ of a configuration $\langle S, R \rangle$ as

$$LB(\langle S, R \rangle) = c_{MST^{w(S \cup L)}(S,R)} - c_{MST^{w(S \cup L)}}.$$

The following proposition proves that $LB(\langle S, R \rangle)$ is indeed a lower bound.

**Proposition 6.** *Let $\langle S, R \rangle$ be an arbitrary configuration and let $L = E \setminus (S \cup R)$. Then, for all $T \in \mathcal{T}(S, R)$, we have*

$$\Delta_T \geq c_{MST^{w(S \cup L)}(S,R)} - c_{MST^{w(S \cup L)}}.$$

*Proof:* Since $T \in \mathcal{T}(S, R)$, it follows that $S \subseteq T \subseteq E - R = S \cup L$. Therefore,

$$c_{MST^{w(T)}} \leq c_{MST^{w(S \cup L)}}$$

On the other hand, by the definition of a minimum spanning tree and since $T \subseteq S \cup L$, we have that

$$c_{MST^{w(S \cup L)}(S,R)} \leq c_T^{w(S \cup L)} = c_T^{w(T)}.$$

The result follows since

$$c_{MST^{w(S \cup L)}(S,R)} - c_{MST^{w(S \cup L)}} \leq c_T^{w(T)} - c_{MST^{w(T)}} = \Delta_T.$$

Observe that this lower bound only requires the computation of two minimum spanning trees and hence it can be computed in $O(m \log m)$ time. Interestingly, $c_{MST^{w(S \cup L)}}$ can use any edge in the graph and is thus independent of the edges selected in $S$ and $R$. Of course, the scenario $w(S \cup L)$ is not! It is easy to see that this lower bound is monotone in both arguments

$$LB(\langle S, R \rangle) \leq LB(\langle S \cup \{e\}, R \rangle)$$
$$LB(\langle S, R \rangle) \leq LB(\langle S, R \cup \{e\} \rangle).$$

## 6 Suboptimality Pruning

A fundamental component of our algorithm is suboptimality pruning, i.e., the ability to remove all non-weak edges at every configuration of the search treee. The results in [Yaman et al., 2001] allows us to detect all weak edges in time $O(m^2 \log m)$ by solving $m$ MSTs. This cost is prohibitive in practice. We now show how to detect all weak edges by solving a single MST and performing a postprocessing step for each edge. The overall complexity is $O(n^2 + m \log m)$, which is $O(m \log m)$ on dense graphs.

*The key idea is to characterize all weak edges in terms of a unique scenario.* The characterization is based on the following propositions which specify when a tree remains an MST under cost changes and the cost of an MST which must contain a specified edge.

**Proposition 7.** *Let $s$ be a scenario and $T \in \mathcal{M}^s$. Let $e = (u,v) \notin T$ and $f = (x,y)$ be the edge of maximal cost on the path from $u$ to $v$. Consider $\hat{s}$ the scenario $s$ where $c_e^s$ is replaced by $c_e^{\hat{s}}$, all other costs remaining the same. Then $T \in \mathcal{M}^{\hat{s}}$ if $c_e^{\hat{s}} \geq c_f^{\hat{s}}$.*

*Proof:* By contradiction. Assume that there exists a tree $T'$ containing $e$ such that $c_{T'}^{\hat{s}} < c_T^{\hat{s}}$. Removing $e$ from $T'$ produces two connected components $C_1$ and $C_2$. If $x \in C_1$ and $y \in C_2$, then we can construct a tree

$$T'' = T' \setminus \{e\} \cup \{f\}$$

and we have

$$c_{T''}^{\hat{s}} = c_{T'}^{\hat{s}} - (c_e^{\hat{s}} - c_f^{\hat{s}}) < c_{T'}^{\hat{s}} < c_T^{\hat{s}}.$$

Since $e \notin T''$ and $e \in T$, we have

$$c_{T''}^s = c_{T''}^{\hat{s}} < c_T^{\hat{s}} = c_T^s$$

which contradicts the fact that $T \in \mathcal{M}^s$. If $x,y \in C_1$ (resp. $C_2$), since there exists a cycle in the graph containing $e$ and $f$, there exists at least one edge $g$ on the path from $u$ to $v$ in $T$ such that $g \in T'$ (otherwise $T'$ would not be a tree). By hypothesis, $c_g^s \leq c_f^s$ and hence $c_e^{\hat{s}} \geq c_g^{\hat{s}}$. We can thus apply the same construction as in the case $x \in C_1$ and $y \in C_2$ with $f$ replaced by $g$.

**Proposition 8.** *Let $s$ be a scenario and $T$ be an $MST^s$. Let $e = (u,v) \notin T$ and $f = (x,y)$ be the edge of maximal cost on the path from $u$ to $v$. Then, $T \setminus \{f\} \cup \{e\} \in \mathcal{M}^s(\{e\}, \emptyset)$.*

```
function MaxCost(u, v vertices) : int
begin
  if u = v then
    return 0;
  else if level(u) < level(v) then
    return max( cost(v,p(v)), MaxCost(u, p(v)));
  else if level(u) > level(v) then
    return max( cost(u,p(u)), MaxCost(p(u), v));
  else
    max_edge = max( cost(u,p(u)), cost(v,p(v)));
    return max( max_edge, MaxCost(p(u), p(v)));
end
```

Figure 3: Finding the Largest Cost of an Edge

The proof of this result is similar to the proof of Proposition 7. We are now ready to present a new characterization of weak edges. The characterization only use the scenario $\bar{s}$ where all costs are at their upper bounds.

**Proposition 9.** *Let $\bar{s}$ be the scenario where all costs are at their upper bounds and $T \in \mathcal{M}^{\bar{s}}$. An edge $e = (u,v)$ is weak if $e \in T$ or if an edge $f$ of maximal cost on the path from $u$ to $v$ in $T$ satisfies $\underline{c_e} \leq \overline{c_f}$.*

*Proof:* Let $\hat{s}$ the scenario $\bar{s}$ where $c_e^{\bar{s}}$ is replaced by $c_e^{\hat{s}}$. If $e \in T$, then $T \in \mathcal{M}^{\hat{s}}$ as well and hence $e$ is weak by Proposition 1. If $e \notin T$ and $\underline{c_e} > \overline{c_f}$, then $T \in \mathcal{M}^{\hat{s}}$ by Proposition 7. By Proposition 8, $T \setminus \{f\} \cup \{e\} \in \mathcal{M}^{\hat{s}}(\{e\}, \emptyset)$ and its cost is greater than $c_T$ since $\underline{c_e} > \overline{c_f}$. Hence $e$ is not weak. If $e \notin T$ and $\underline{c_e} = \overline{c_f}$, then $T \setminus \{f\} \cup \{e\}$ is an $MST^{\hat{s}}$ and hence $e$ is weak. The same holds for the case where $e \notin T$ and $\underline{c_e} < \overline{c_f}$.

We now describe how to use Proposition 9 to obtain an $O(m \log m)$ algorithm on dense graphs. The key idea is to compute $MST^{\bar{s}}$, i.e., the MST when all costs are at their upper bounds. All edges in this MST are weak. In addition, for each $e = (u, v)$ not in the MST, we compute the largest cost of any edge on the path from $u$ to $v$. This can be done easily by upgrading Prim's algorithm slightly to associate a level and a parent to each vertex. When an edge $e = (u, v)$ with $u \in T$ and $v \notin T$ is added to $T$ in Prim's algorithm, we set

```
level(v) = level(u) + 1;
parent(v) = u;
```

These modifications do not affect the complexity of the algorithm. It now suffices to apply the algorithm depicted in Figure 3 to compute the cost of the maximal edge and to apply Proposition 9. The algorithm in Figure 3 simply follows the paths from $u$ and $v$ to their common ancestor, computing the cost of the maximal edge on the way. Since algorithm MaxCost takes $O(n)$ in the worst case, the overall complexity becomes $O(m \log m + mn)$. However, it is easy to reduce this complexity to $O(m \log m + n^2)$ by amortizing the computation of the maximal costs. It suffices to cache the results of MaxCost in an $n \times n$ matrix $M$. For each edge $e = (u,v) \in E \setminus T$ we first examine entry $M[u,v]$ and call MaxCost$(u,v)$ only if this entry is uninitialized. Since there are at most $n^2$ entries and each call to MaxCost$(u,v)$ costs $O(1)$ per entry it fills, the overall complexity complexity becomes $O(m \log m + n^2)$. We proved the following theorem.

**Theorem 1.** *All weak edges of a graph can be computed in $O(m \log m + n^2)$ time.*

## 7 Branching Strategy

It is well-known that a good branching heuristic may improve performance significantly. We now show a branching heuristic adapting the first-fail principle [Haralick and Elliot, 1980] to robust optimization. The key idea is to explore the most uncertain edges first, i.e., to branch on an edge $e$ with the maximal difference $\overline{c_e} - \underline{c_e}$. Indeed, rejecting $e$ (i.e., adding $e$ to $R$) allows $MST^{w(S \cup L)}$ to select $e$ at a low cost, possibly giving a large deviation. Hence this branch is likely to fail early. However, this is only important if $MST^{w(S \cup L)}$ is likely to select $e$ which may not necessarily the case if $\underline{c_e}$ is large compared to other "similar" edges. Hence, it seems appropriate to select first an edge $e \in MST^{w(S \cup L)}$ whose difference $\overline{c_e} - \underline{c_e}$ is maximal. This justifies the following branching strategy.

**Definition 7 (Branching Strategy).** *Let $\langle S, R \rangle$ be a configuration, $L^* = L \cap MST^{w(S \cup L)}$, and $L^- = L \setminus MST^{w(S \cup L)}$. The branching strategy selects an edge $s$ defined as follows:*

$$s = \begin{cases} \arg\text{-}\max_{e \in L^*} \overline{c_e} - \underline{c_e} & \text{if } L^* \neq \emptyset \\ \arg\text{-}\max_{e \in L^-} \overline{c_e} - \underline{c_e} & \text{otherwise.} \end{cases}$$

## 8 Experimental Results

We now report experimental results comparing the constraint atisfaction and the MIP approaches. The comparison uses the instances in [Yaman et al., 2001], as well as some new instances which capture additional structure arising in practical applications.

The experimental setting of [Yaman et al., 2001] uses complete graphs with $n$ vertices and six classes of problems. Three of the six classes use tight intervals for edge costs, while the other three allowed larger differences between the lower and upper bounds. The edge intervals were chosen as follows. The values of $\underline{c_e}$ and $\overline{c_e}$ are uniformly distributed in the intervals:

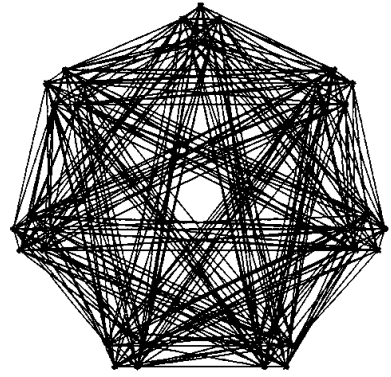

Figure 4: A Class 7 Network

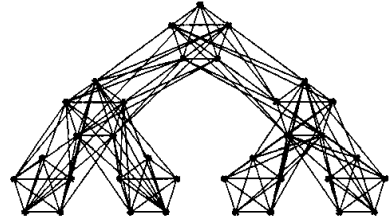

Figure 5: A Class 8 Network

class 1:　$\underline{c_e} \in [0, 10]$, $\overline{c_e} \in (\underline{c_e}, 10]$
class 2:　$\underline{c_e} \in [0, 15]$, $\overline{c_e} \in (\underline{c_e}, 15]$
class 3:　$\underline{c_e} \in [0, 20]$, $\overline{c_e} \in (\underline{c_e}, 20]$
class 4:　$\underline{c_e} \in [0, 10]$, $\overline{c_e} \in (\underline{c_e}, 20]$
class 5:　$\underline{c_e} \in [0, 15]$, $\overline{c_e} \in (\underline{c_e}, 30]$
class 6:　$\underline{c_e} \in [0, 20]$, $\overline{c_e} \in (\underline{c_e}, 40]$

Note that the size of the search space to explore is $O(2^{300})$ for a complete graph of 25 nodes. Of course, our constraint satisfaction algorithm will only explore a small fraction of that space. In addition to these six classes, we also generate instances (Classes 7 and 8) whose cost structure is not the same for all the edges. Class 7 contains instances which represent a two-level network. The lower level consists of clusters of 5 nodes whose edges are generated according to Class 1 above. The upper level links the clusters and these edges have higher costs, i.e., Class 1 costs shifted by a constant which is larger than the Class 1 edges. This captures the fact that, in networks, there are often various types of edges with different costs. See Figure 4 for an instance of Class 7 with 35 nodes. Class 8 contains instances which are similar to Class 7, except that the upper-level layer is organized as a binary tree. See Figure 5 for an instance of Class 8 with 35 nodes. In general, classes 7 and 8 are significantly harder than classes 1 to 6, since preprocessing is less effective than in Classes 1 to 6 because of the additional structure. Observe also that these instances are sparser for the same number of nodes.

Table 2 compares the efficiency of the two approaches.

| Edges(Nodes) | Algo. | Class 1 | Class 2 | Class 3 | Class 4 | Class 5 | Class 6 | Class 7 | Class 8 |
|---|---|---|---|---|---|---|---|---|---|
| 45(10) | CSR | 0.12 | 0.07 | 0.08 | 0.08 | 0.06 | 0.07 | 0.06 | 0.06 |
| | CSF | 0.14 | 0.10 | 0.12 | 0.12 | 0.09 | 0.11 | 0.07 | 0.07 |
| | MIP | 6.59 | 3.95 | 4.10 | 4.38 | 4.68 | 3.83 | 2.29 | 2.02 |
| 105(15) | CSR | 1.82 | 1.96 | 1.09 | 0.36 | 0.86 | 0.33 | 1.76 | 1.15 |
| | CSF | 2.53 | 3.06 | 2.15 | 0.86 | 2.01 | 0.77 | 1.83 | 1.35 |
| | MIP | 245.19 | 184.14 | 136.88 | 109.30 | 91.33 | 87.62 | 730.20 | 88.66 |
| 190(20) | CSR | 39.72 | 33.80 | 8.91 | 3.43 | 2.09 | 3.12 | 7.37 | 9.51 |
| | CSF | 74.86 | 61.08 | 12.66 | 7.56 | 4.69 | 7.05 | 5.28 | 6.77 |
| | MIP | 8620.66 | 5517.75 | 14385.55 | 3344.95 | 12862.29 | 22855.82 | 18547.27 | 1399.48 |
| 300(25) | CSR | 121.85 | 181.57 | 68.42 | 20.41 | 12.23 | 13.26 | 91.89 | 61.71 |
| | CSF | 244.18 | 272.63 | 145.11 | 50.37 | 30.55 | 32.41 | 97.94 | 36.21 |
| 435(30) | CSR | 926.65 | 415.07 | 942.38 | 133.90 | 63.85 | 177.47 | 1719.61 | 721.28 |
| | CSF | 2100.43 | 909.15 | 1811.88 | 359.72 | 167.58 | 418.41 | 804.78 | 284.39 |
| 595(35) | CSR | 4639.55 | 5095.71 | 2304.36 | 383.34 | 188.36 | 419.56 | 32511.77 | 6356.78 |
| | CSF | 10771.37 | 11906.01 | 4183.22 | 1100.88 | 548.01 | 1115.83 | 14585.73 | 2149.58 |
| 780(40) | CSR | 27206.38 | 16388.12 | 15059.39 | 1103.50 | 1122.57 | 1071.62 | 57309.23 | 33390.67 |
| | CSF | 29421.70 | 34666.84 | 22084.00 | 3456.84 | 3241.04 | 3031.55 | 28432.91 | 11339.59 |

Table 2: Average CPU Time of the Algorithms

It reports the computation times in CPU seconds of the MIP implementations, the constraint satisfaction algorithm with suboptimality pruning at the root node only (CSR), and the constraint satisfaction algorithm with suboptimality pruning at every node (CSF). The times are given on a Sun Sparc 440Mhz processor and the average is computed over 10 instances for each class and each size. We used CPLEX 6.51 for solving the MIP, after preprocessing of the weak and strong edges.

*Observe that the constraint satisfaction approach produces extremely dramatic speedups over the MIP approach.* On Class 1, CSR runs about 134 times faster than the MIP on graphs with 15 nodes and about 217 times faster on graphs with 20 nodes. On Classes 7 and 8, the speed-ups are even more impressive. On graphs with 20 nodes for classes 7 and 8, CSR runs about 3500 and 200 times faster than the MIP. (Recall that these graphs are not complete). The MIP times are not given for graphs with more than 20 nodes, since they cannot be obtained in reasonable time. The results indicate that the constraint satisfaction approach is also better at exploiting the network structure, which is likely to be fundamental in practice. Observe also that the constraint satisfaction approach is able to tackle much large instances in reasonable times.

Figures 6, 7, and 8 plot the execution times of the two constraint satisfaction algorithms. Observe also that applying suboptimality pruning at every node is not cost-effective on Classes 1 to 6. This is due to the fact that the graphs are complete and the costs are uniformly distributed in these instances. Classes 7 and 8, which add a simple additional cost structure, clearly indicate that suboptimality pruning becomes increasingly important when the network is more heterogeneous. The benefits of suboptimality pruning clearly appears on large graphs for class 8, where CSF is about three times as fast in the average. In fact, CSF is sig-

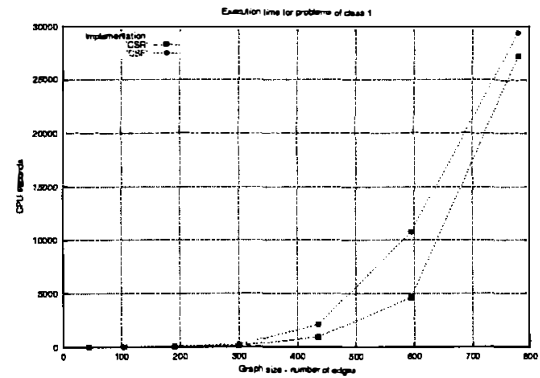

Figure 6: Constraint Satisfaction on Class 1.

nificantly faster (e.g., 10 times faster) than CSR on some instances, while the two algorithms are similar on others. *It is important to mention that systematic suboptimality pruning is useless without Theorem 1. Indeed, the pruning benefit is often offset by the high pruning cost otherwise.*

Overall, it is clear that the constraint satisfaction approach is much more effective on these problems than the MIP approach. It produces extremely dramatic speed-ups and substantially enlarge the class of instances that are amenable to effective solutions. The constraint satisfaction approach is able to solve large-scale problems (over 1,000 edges). Since networks are often organized in hierarchies, it should scale up nicely to larger real-life instances. There is still considerable room for improvement in the implementation, since incremental MST algorithms [Rauch Henzinger and King, 1997] and enhanced feasibility pruning may decrease runtime significantly.

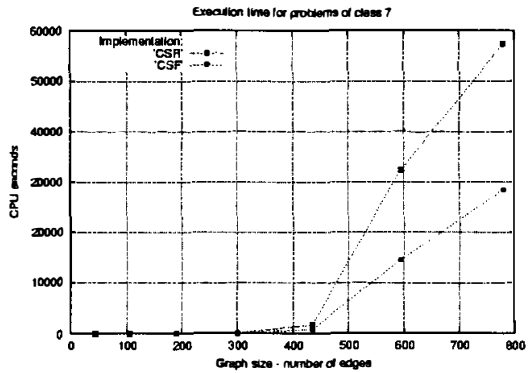

Figure 7: Constraint Satisfaction on Class 7.

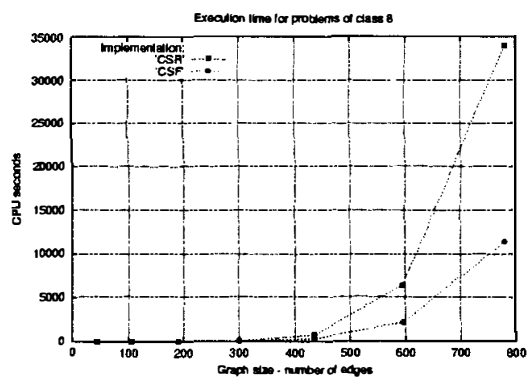

Figure 8: Constraint Satisfaction on Class 8.

## 9 Conclusion

This paper reconsidered the robust minimum spanning tree with interval data, which has received much attention in recent years due to its importance in communication networks. It applied a constraint satisfaction approach to the RSTPIE and proposed a search algorithm based on three important and orthogonal components: (1) a combinatorial lower bound to eliminate suboptimal trees; (2) a pruning component which eliminates edges that cannot be part of any feasible or optimal solution; and (3) a branching heuristic which selects the most uncertain edges first. The lower bound exploits the combinatorial structure of the problem and reduces to solving two MST problems. The pruning component eliminates infeasible edges (e.g., edges that would lead to cycles and non-connected components) as well as suboptimal edges (i.e., edges that cannot appear in optimal solutions). It uses the concept of weak edges and a new $O(m \log m)$ amortized algorithm for detecting all weak edges (where $m$ is the size of the graph), improving the results of Yaman & al by an order of magnitude. The overall algorithm provides very dramatic speed-ups over the MIP approach on the 15 and 20 node graphs proposed in [Yaman et al., 2001]. On graphs with additional cost structure, the benefits are even more impressive. For instance, the constraint satisfaction algorithm is about 3,500 times faster on Class 7 graphs with 20 nodes. The constraint satisfaction approach also solved large-scale instances with up to 50 nodes (725 edges).

There are many open issues arising from this research. On the one hand, it would be interesting to find an $O(m \log m)$ algorithm for detecting strong edges, to quantify the benefits of incremental MST algorithms, to investigate more sophisticated feasibility pruning, and to evaluate the approach on sparser graphs. Finally, it would be interesting to study whether suboptimal conditions can be derived for a wide variety of (robust) combinatorial optimization problems given the ubiquity of these problems in practice. In particular, robust matching and robust shortest paths have fundamental applications in computational finance and networks and are worth investigating.

## Acknowledgments

Thanks to the reviewers for their detailed comments. Ionut Aron is supported by NSF ITR Award ACI-0121497. Pascal Van Hentenryck is partially supported by NSF ITR Award DMI-0121495.